\documentclass[11pt,a4paper]{article}
\usepackage[hyperref]{acl2018}
\usepackage{times}
\usepackage{latexsym}

\usepackage{url}

\usepackage{wrapfig,lipsum,booktabs}
\usepackage{amsmath}
\usepackage{amssymb}
\usepackage{subfig}
\usepackage{adjustbox} 
\usepackage{caption}

\usepackage{times}
\usepackage{graphicx}
\usepackage{amsmath,amssymb,enumerate,bbm,color,alltt}
\usepackage{graphicx}
\usepackage{float}
\usepackage{stfloats}
\usepackage{subfig}
\usepackage{longtable}
\usepackage{tabularx}
\usepackage{multirow}
\usepackage{booktabs}
\usepackage{subfig}
\usepackage{hyperref}

\aclfinalcopy 
\def\aclpaperid{1018} 

\title{Baseline Needs More Love: On Simple Word-Embedding-Based Models and Associated Pooling Mechanisms}

\author{Dinghan Shen$^{\mathbf{1}}$, ~Guoyin Wang$^{\mathbf{1}}$, ~Wenlin Wang$^{\mathbf{1}}$, ~	Martin Renqiang Min$^{\mathbf{2}}$ 
	\smallskip 
	~  \\
	\bf{Qinliang Su$^{\mathbf{3}}$, ~Yizhe Zhang$^{\mathbf{4}}$, ~Chunyuan Li$^{\mathbf{1}}$, ~Ricardo Henao$^{\mathbf{1}}$, ~Lawrence Carin$^{\mathbf{1}}$} \\
	\smallskip 
	$^{\mathbf{1}}$ Duke University~~~~
	$^{\mathbf{2}}$ NEC Laboratories America~~~~
	$^{\mathbf{3}}$ Sun Yat-sen University~~~~
	$^{\mathbf{4}}$ Microsoft Research \\
	{\tt dinghan.shen@duke.edu} 
	}

\date{}

\begin{document}

\maketitle

\begin{abstract}
Many deep learning architectures have been proposed to model the \emph{compositionality} in text sequences, requiring a substantial number of parameters and expensive computations.
However, there has not been a rigorous evaluation regarding the added value of sophisticated compositional functions.
In this paper, we conduct a point-by-point comparative study between Simple Word-Embedding-based Models (SWEMs), consisting of parameter-free pooling operations, relative to word-embedding-based RNN/CNN models.
Surprisingly, SWEMs exhibit comparable or even superior performance in the majority of cases considered.
Based upon this understanding, we propose two additional pooling strategies over learned word embeddings: ($i$) a max-pooling operation for improved interpretability; and ($ii$) a hierarchical pooling operation, which preserves spatial ($n$-gram) information within text sequences.
We present experiments on 17 datasets encompassing three tasks: ($i$) (long) document classification; ($ii$) text sequence matching; and ($iii$) short text tasks, including classification and tagging.
The source code and datasets can be obtained from \url{https://github.com/dinghanshen/SWEM}.
\end{abstract}
\section{Introduction}\label{introduction}
Word embeddings, learned from massive unstructured text data, are widely-adopted building blocks for Natural Language Processing (NLP).
By representing each word as a fixed-length vector, these embeddings can group semantically similar words, while implicitly encoding rich linguistic regularities and patterns \citep{bengio2003neural, mikolov2013distributed, pennington2014glove}.
Leveraging the word-embedding construct, many deep architectures have been proposed to model the \emph{compositionality} in variable-length text sequences.
These methods range from simple operations like addition \citep{mitchell2010composition, iyyer2015deep}, to more sophisticated compositional functions such as Recurrent Neural Networks (RNNs) \citep{tai2015improved, sutskever2014sequence}, Convolutional Neural Networks (CNNs) \citep{kalchbrenner2014convolutional, kim2014convolutional, Zhang2017AdversarialFM} and Recursive Neural Networks \citep{socher2011parsing}.

Models with more expressive compositional functions, \emph{e.g.}, RNNs or CNNs, have demonstrated impressive results; however, they are typically computationally expensive, due to the need to estimate hundreds of thousands, if not millions, of parameters \citep{parikh2016decomposable}.
In contrast, models with simple compositional functions often compute a sentence or document embedding by simply adding, or averaging, over the word embedding of each sequence element obtained via, \emph{e.g.}, \emph{word2vec} \citep{mikolov2013distributed}, or \emph{GloVe} \citep{pennington2014glove}.
Generally, such a Simple Word-Embedding-based Model (SWEM) does not explicitly account for spatial, word-order information within a text sequence.
However, they possess the desirable property of having significantly fewer parameters, enjoying much faster training, relative to RNN- or CNN-based models.
Hence, there is a computation-\emph{vs.}-expressiveness tradeoff regarding how to model the compositionality of a text sequence.

In this paper, we conduct an extensive experimental investigation to understand when, and why, simple pooling strategies, operated over word embeddings alone, already carry sufficient information for natural language understanding.
To account for the distinct nature of various NLP tasks that may require different semantic features, we compare SWEM-based models with existing recurrent and convolutional networks in a point-by-point manner.
Specifically, we consider 17 datasets, including three distinct NLP tasks: \emph{document classification} (Yahoo news, Yelp reviews, \emph{etc}.), \emph{natural language sequence matching} (SNLI, WikiQA, \emph{etc}.) and \emph{(short) sentence classification/tagging} (Stanford sentiment treebank, TREC, \emph{etc}.). Surprisingly, SWEMs exhibit comparable or even superior performance in the majority of cases considered.

In order to validate our experimental findings, we conduct additional investigations to understand to what extent \emph{the word-order information} is utilized/required to make predictions on different tasks. We observe that in text representation tasks, many words (\emph{e.g.}, stop words, or words that are not related to sentiment or topic) do not meaningfully contribute to the final predictions (\emph{e.g.}, sentiment label). Based upon this understanding, we propose to leverage a \emph{max-pooling} operation directly over the word embedding matrix of a given sequence, to select its most \emph{salient} features.
This strategy is demonstrated to extract complementary features relative to the standard averaging operation, while resulting in a more interpretable model.
Inspired by a case study on sentiment analysis tasks, we further propose a \emph{hierarchical pooling} strategy to abstract and preserve the spatial information in the final representations.
This strategy is demonstrated to exhibit comparable empirical results to LSTM and CNN on tasks that are sensitive to word-order features, while maintaining the favorable properties of not having compositional parameters, thus fast training.

Our work presents a simple yet strong baseline for text representation learning that is widely ignored in benchmarks, and highlights the general computation-\emph{vs.}-expressiveness tradeoff associated with appropriately selecting compositional functions for distinct NLP problems.
Furthermore, we quantitatively show that the word-embedding-based text classification tasks can have the similar level of difficulty regardless of the employed models, using the subspace training~\cite{li_id_2018_ICLR} to constrain the trainable parameters. Thus, according to Occam's razor, simple models are preferred.

\section{Related Work}\label{related_work}
A fundamental goal in NLP is to develop expressive, yet computationally efficient compositional functions that can capture the linguistic structure of natural language sequences.
Recently, several studies have suggested that on certain NLP applications, much simpler word-embedding-based architectures exhibit comparable or even superior performance, compared with more-sophisticated models using recurrence or convolutions \citep{parikh2016decomposable, vaswani2017attention}.
Although complex compositional functions are avoided in these models, additional modules, such as attention layers, are employed on top of the word embedding layer.
As a result, the specific role that the word embedding plays in these models is not emphasized (or explicit), which distracts from understanding how important the word embeddings alone are to the observed superior performance.
Moreover, several recent studies have shown empirically that the advantages of distinct compositional functions are highly dependent on the specific task \citep{mitchell2010composition, iyyer2015deep, zhang2015fixed, wieting2015towards, arora2016simple}.
Therefore, it is of interest to study the practical value of the additional expressiveness, on a wide variety of NLP problems.

SWEMs bear close resemblance to Deep Averaging Network (DAN) \citep{iyyer2015deep} or fastText \citep{joulin2016bag}, where they show that average pooling achieves promising results on certain NLP tasks.
However, there exist several key differences that make our work unique. First, we explore a series of pooling operations, rather than only average-pooling.
Specifically, a \emph{hierarchical} pooling operation is introduced to incorporate spatial information, which demonstrates superior results on sentiment analysis, relative to average pooling.
Second, our work not only explores when simple pooling operations are enough, but also investigates the underlying reasons, \emph{i.e.}, what semantic features are required for distinct NLP problems.
Third, DAN and fastText only focused on one or two problems at a time, thus a comprehensive study regarding the effectiveness of various compositional functions on distinct NLP tasks, \emph{e.g.}, categorizing short sentence/long documents, matching natural language sentences, has heretofore been absent.
In response, our work seeks to perform a comprehensive comparison with respect to simple-\emph{vs.}-complex compositional functions, across a wide range of NLP problems, and reveals some general rules for rationally selecting models to tackle different tasks.

\section{Models \& training}\label{model}
\vspace{-1mm}
Consider a text sequence represented as $X$ (either a sentence or a document), composed of a sequence of words: $\{w_1, w_2, ...., w_L\}$, where $L$ is the number of tokens, \emph{i.e.}, the sentence/document length.
Let $\{v_1, v_2, ...., v_L\}$ denote the respective word embeddings for each token, where $v_l\in\mathbb{R}^K$.
The compositional function, $X \to z$, aims to combine word embeddings into a fixed-length sentence/document representation $z$.
These representations are then used to make predictions about sequence $X$.
Below, we describe different types of functions considered in this work.

\subsection{Recurrent Sequence Encoder}\label{rnn}
\vspace{-1mm}
A widely adopted compositional function is defined in a recurrent manner: the model successively takes word vector $v_t$ at position $t$, along with the hidden unit $h_{t-1}$ from the last position $t-1$, to update the current hidden unit via $h_t = f(v_t, h_{t-1})$, where $f(\cdot)$ is the transition function.

To address the issue of learning long-term dependencies, $f(\cdot)$ is often defined as Long Short-Term Memory (LSTM) \citep{hochreiter1997long}, which employs \emph{gates} to control the flow of information abstracted from a sequence.
We omit the details of the LSTM and refer the interested readers to the work by \citet{graves2013hybrid} for further explanation.
Intuitively, the LSTM encodes a text sequence considering its word-order information, but yields additional compositional parameters that must be learned.

\subsection{Convolutional Sequence Encoder}\label{cnn}
The Convolutional Neural Network (CNN) architecture \citep{kim2014convolutional, collobert2011natural, gan2017learning, zhang2017deconvolutional, shen2017deconvolutional} is another strategy extensively employed as the compositional function to encode text sequences.
The convolution operation considers windows of $n$ consecutive words within the sequence, where a set of filters (to be learned) are applied to these word windows to generate corresponding \emph{feature maps}.
Subsequently, an aggregation operation (such as max-pooling) is used on top of the feature maps to abstract the most salient semantic features, resulting in the final representation.
For most experiments, we consider a single-layer CNN text model.
However, Deep CNN text models have also been developed \citep{conneau2016very}, and are considered in a few of our experiments.

\subsection{Simple Word-Embedding Model (SWEM)}\label{swem}
\vspace{-1mm}
To investigate the raw modeling capacity of word embeddings, we consider a class of models with no additional compositional parameters to encode natural language sequences, termed SWEMs.
Among them, the simplest strategy is to compute the element-wise average over word vectors for a given sequence \cite{wieting2015towards, adi2016fine}:
\vspace{-2mm}
\begin{align}\label{eq:ave}
z = \frac{1}{L} \sum_{i=1}^{L} v_i \,.
\end{align}
%
The model in \eqref{eq:ave} can be seen as an average pooling operation, which takes the mean over each of the $K$ dimensions for all word embeddings, resulting in a representation $z$ with the same dimension as the embedding itself, termed here SWEM-\emph{aver}.
Intuitively, $z$ takes the information of every sequence element into account via the addition operation.

\paragraph{Max Pooling}
Motivated by the observation that, in general, only a small number of key words contribute to final predictions, we propose another SWEM variant, that extracts the most salient features from every word-embedding dimension, by taking the maximum value along each dimension of the word vectors.
This strategy is similar to the max-over-time pooling operation in convolutional neural networks \citep{collobert2011natural}:
\begin{align}\label{eq:max}
z =\textbf{\normalfont{Max-pooling}}(v_1, v_2, ..., v_L) \,.
\end{align}
We denote this model variant as SWEM-\emph{max}.
Here the $j$-th component of $z$ is the maximum element in the set $\{v_{1j},\dots,v_{Lj}\}$, where $v_{1j}$ is, for example, the $j$-th component of $v_1$. With this pooling operation, those words that are unimportant or unrelated to the corresponding tasks will be ignored in the encoding process (as the components of the embedding vectors will have small amplitude), unlike SWEM-\emph{aver} where every word contributes equally to the representation.

Considering that SWEM-\emph{aver} and SWEM-\emph{max} are complementary, in the sense of accounting for different types of information from text sequences, we also propose a third SWEM variant, where the two abstracted features are concatenated together to form the sentence embeddings, denoted here as SWEM-\emph{concat}.
For all SWEM variants, there are no additional compositional parameters to be learned.
As a result, the models only exploit intrinsic word embedding information for predictions.

\paragraph{Hierarchical Pooling}\label{swem_lg}
Both SWEM-\emph{aver} and SWEM-\emph{max} do not take word-order or spatial information into consideration, which could be useful for certain NLP applications.
So motivated, we further propose a \emph{hierarchical} pooling layer.
Let $v_{i:i+n-1}$ refer to the \emph{local} window consisting of $n$ consecutive words words, $v_i , v_{i+1}, . . . , v_{i+n-1}$.
First, an average-pooling is performed on each local window, $v_{i:i+n-1}$.
The extracted features from all windows are further down-sampled with a \emph{global} max-pooling operation on top of the representations for every window.
We call this approach SWEM-\emph{hier} due to its layered pooling.

This strategy preserves the local spatial information of a text sequence in the sense that it keeps track of how the sentence/document is constructed from individual word windows, \emph{i.e.}, $n$-grams.
This formulation is related to bag-of-$n$-grams method \cite{zhang2015character}. 
However, SWEM-\emph{hier} learns fixed-length representations for the $n$-grams that appear in the corpus, rather than just capturing their occurrences via count features, which may potentially advantageous for prediction purposes.

\begin{table}[t!]
	\def\arraystretch{1.0}
	\resizebox{\columnwidth}{!}{%
		\begin{tabular} {c||c|c|c}    
			\toprule[1.2pt]
			\textbf{Model} & Parameters & Complexity & Sequential Ops \\
			\hline
			CNN    & $n\cdot K \cdot d$ & $\mathcal{O}(n\cdot L\cdot K \cdot d)$ & $\mathcal{O}(1)$  \\
			LSTM   & $4 \cdot d\cdot (K+d)$ & $\mathcal{O}(L\cdot d^2 + L \cdot K \cdot d)$ & $\mathcal{O}(L)$   \\
			SWEM         & 0 & $\mathcal{O}(L\cdot K)$ &  $\mathcal{O}(1)$  \\
			\bottomrule[1.2pt]
		\end{tabular}
	}
	\caption{Comparisons of CNN, LSTM and SWEM architectures. Columns correspond to the number of \emph{compositional} parameters, computational complexity and sequential operations, respectively.}
	\label{tab:comparison}
	\vspace{-1mm}
\end{table}

\begin{table*}[t!]
	\centering
	\def\arraystretch{1.0}
	\begin{small}
	\begin{tabular}{c||c|c|c|c|c}
		\toprule[1.2pt]
		\textbf{Model} &  	\textbf{Yahoo! Ans.} & \textbf{AG News} & 	\textbf{Yelp P.} & \textbf{Yelp F.} & \textbf{DBpedia} \\
		\hline
		Bag-of-means$^{\ast}$        & 60.55 & 83.09 & 87.33  & 53.54  &  90.45  \\
		Small word CNN$^{\ast}$        & 69.98 & 89.13 & 94.46 & 58.59 & 98.15  \\ 
		Large word CNN$^{\ast}$        & 70.94 & 91.45 & 95.11 & 59.48  & 98.28  \\ 
		LSTM$^{\ast}$        & 70.84 & 86.06 & 94.74 & 58.17 & 98.55  \\ 
		Deep CNN (29 layer)$^{\dagger}$       & 73.43 & 91.27 & \bf{95.72} &\bf{64.26} &  \bf{98.71} \\
		fastText $^{\ddagger}$  & 72.0 & 91.5 &  93.8 & 60.4 &  98.1 \\
		fastText (bigram)$^{\ddagger}$   & 72.3 & 92.5 & 95.7 & 63.9 &  98.6 \\
		\hline
		SWEM-\emph{aver}          & 73.14 & 91.71 & 93.59 & 60.66 & 98.42 \\
		SWEM-\emph{max}         & 72.66  & 91.79 & 93.25 & 59.63 & 98.24  \\
		SWEM-\emph{concat}        & \bf{73.53}  & \bf{92.66} & 93.76 & 61.11 & \bf{98.57}  \\
		\hline
		SWEM-\emph{hier}    & 73.48  & 92.48 & \bf{95.81} & \bf{63.79} & 98.54  \\
		\bottomrule[1.2pt]
	\end{tabular}
	\end{small}
	\vspace{-2mm}
	\caption{Test accuracy on (long) document classification tasks, in percentage. Results marked with $\ast$ are reported in \citet{zhang2015character}, with $\dagger$ are reported in \citet{conneau2016very}, and with $\ddagger$ are reported in \citet{joulin2016bag}.}
	\label{tab:document}
	\vspace{0mm}
\end{table*}

\subsection{Parameters \& Computation Comparison}\label{compare}
\vspace{-1mm}
We compare CNN, LSTM and SWEM wrt their parameters and computational speed.
$K$ denotes the dimension of word embeddings, as above.
For the CNN, we use $n$ to denote the filter width (assumed constant for all filters, for simplicity of analysis, but in practice variable $n$ is commonly used).
We define $d$ as the dimension of the final sequence representation.
Specifically, $d$ represents the dimension of hidden units or the number of filters in LSTM or CNN, respectively.

We first examine the number of \emph{compositional parameters} for each model.
As shown in Table~\ref{tab:comparison}, both the CNN and LSTM have a large number of parameters, to model the semantic compositionality of text sequences, whereas SWEM has no such parameters.
Similar to \citet{vaswani2017attention}, we then consider the computational complexity and the minimum number of sequential operations required for each model.
SWEM tends to be more efficient than CNN and LSTM in terms of computation complexity.
For example, considering the case where $K=d$, SWEM is faster than CNN or LSTM by a factor of $nd$ or $d$, respectively.
Further, the computations in SWEM are highly parallelizable, unlike LSTM that requires $\mathcal{O}(L)$ sequential steps.

\section{Experiments}\label{experiments}
\vspace{-1mm}
We evaluate different compositional functions on a wide variety of supervised tasks, including document categorization, text sequence matching (given a sentence pair, $X_1$, $X_2$, predict their relationship, $y$) as well as (short) sentence classification.
We experiment on 17 datasets concerning natural language understanding, with corresponding data statistics summarized in the Supplementary Material.

We use GloVe word embeddings with $K=300$ \citep{pennington2014glove} as initialization for all our models. 
Out-Of-Vocabulary (OOV) words are initialized from a uniform distribution with range $[-0.01, 0.01]$.
The GloVe embeddings are employed in two ways to learn refined word embeddings: ($i$) directly updating each word embedding during training; and ($ii$) training a 300-dimensional Multilayer Perceptron (MLP) layer with ReLU activation, with GloVe embeddings as input to the MLP and with output defining the refined word embeddings.
The latter approach corresponds to learning an MLP model that adapts GloVe embeddings to the dataset and task of interest.
The advantages of these two methods differ from dataset to dataset.
We choose the better strategy based on their corresponding performances on the validation set.
The final classifier is implemented as an MLP layer with dimension selected from the set $[100, 300, 500, 1000]$, followed by a sigmoid or softmax function, depending on the specific task.

Adam \citep{kingma2014adam} is used to optimize all models, with learning rate selected from the set $[1 \times 10^{-3}, 3 \times 10^{-4}, 2 \times 10^{-4}, 1 \times 10^{-5}]$ (with cross-validation used to select the appropriate parameter for a given dataset and task).
Dropout regularization \citep{srivastava2014dropout} is employed on the word embedding layer and final MLP layer, with dropout rate selected from the set $[0.2, 0.5, 0.7]$.
The batch size is selected from $[2, 8, 32, 128, 512]$.

\begin{table*}[t!]
	\centering 
	\def\arraystretch{1.1}
	\begin{small}
		\begin{tabular}{c|c|c|c|c|c|c}
			\toprule[1.2pt]
			\textbf{Politics} & \textbf{Science} & \textbf{Computer} & \textbf{Sports} & \textbf{Chemistry} & \textbf{Finance} & \textbf{Geoscience}  \\
			\hline
			philipdru & coulomb & system32 & billups & sio2 (SiO$_2$) & proprietorship & fossil  \\
			justices & differentiable & cobol & midfield & nonmetal & ameritrade & zoos \\
			impeached & paranormal & agp & sportblogs & pka & retailing & farming \\
			impeachment & converge & dhcp & mickelson  & chemistry & mlm & volcanic  \\
			neocons & antimatter & win98 & juventus & quarks & budgeting & ecosystem  \\
			\bottomrule[1.2pt]
		\end{tabular}
		\caption{Top five words with the largest values in a given word-embedding dimension (each column corresponds to a dimension). The first row shows the (manually assigned) topic for words in each column.}
		\label{tab:similar}
	\end{small}
	\vspace{-3mm}
\end{table*}

\subsection{Document Categorization}
We begin with the task of categorizing documents (with approximately 100 words in average per document).
We follow the data split in \citet{zhang2015character} for comparability.
These datasets can be generally categorized into three types: \emph{topic categorization} (represented by Yahoo! Answer and AG news), \emph{sentiment analysis} (represented by Yelp Polarity and Yelp Full) and \emph{ontology classification} (represented by DBpedia).
Results are shown in Table~\ref{tab:document}.
Surprisingly, on topic prediction tasks, our SWEM model exhibits stronger performances, relative to both LSTM and CNN compositional architectures, this by leveraging both the average and max-pooling features from word embeddings.
Specifically, our SWEM-\emph{concat} model even outperforms a 29-layer deep CNN model \citep{conneau2016very}, when predicting topics.
On the ontology classification problem (DBpedia dataset), we observe the same trend, that SWEM exhibits comparable or even superior results, relative to CNN or LSTM models.

Since there are no compositional parameters in SWEM, our models have an order of magnitude fewer parameters (excluding embeddings) than LSTM or CNN, and are considerably more computationally efficient.
As illustrated in Table~\ref{tab:yahoo}, SWEM-\emph{concat} achieves better results on Yahoo! Answer than CNN/LSTM, with only 61K parameters (one-tenth the number of LSTM parameters, or one-third the number of CNN parameters), while taking a fraction of the training time relative to the CNN or LSTM.

\begin{table}[h!]
	\centering 
	\def\arraystretch{1.0}
	\begin{small}
		\begin{tabular} {c||c|c} 
			\toprule[1.2pt]
			\textbf{Model} & Parameters & Speed \\
			\hline
			CNN    & 541K & 171s   \\
			LSTM  & 1.8M & 598s   \\
			SWEM     & \textbf{61K} &  \textbf{63s} \\
			\bottomrule[1.2pt]
		\end{tabular}
	\end{small}
	\caption{Speed \& Parameters on Yahoo! Answer dataset.}
	\label{tab:yahoo}
	\vspace{-5mm}
\end{table}

Interestingly, for the sentiment analysis tasks, both CNN and LSTM compositional functions perform better than SWEM, suggesting that word-order information may be required for analyzing sentiment orientations.
This finding is consistent with \citet{pang2002thumbs}, where they hypothesize that the positional information of a word in text sequences may be beneficial to predict sentiment.
This is intuitively reasonable since, for instance, the phrase ``not really good'' and ``really not good'' convey different levels of negative sentiment, while being different only by their word orderings.
Contrary to SWEM, CNN and LSTM models can both capture this type of information via convolutional filters or recurrent transition functions.
However, as suggested above, such word-order patterns may be much less useful for predicting the topic of a document.
This may be attributed to the fact that word embeddings alone already provide sufficient topic information of a document, at least when the text sequences considered are relatively long.
\subsubsection{Interpreting model predictions}
\begin{figure}[h!] 
	\centering 
	\def\arraystretch{1.0}
	\vspace{-2mm}
	\includegraphics[scale=0.3]{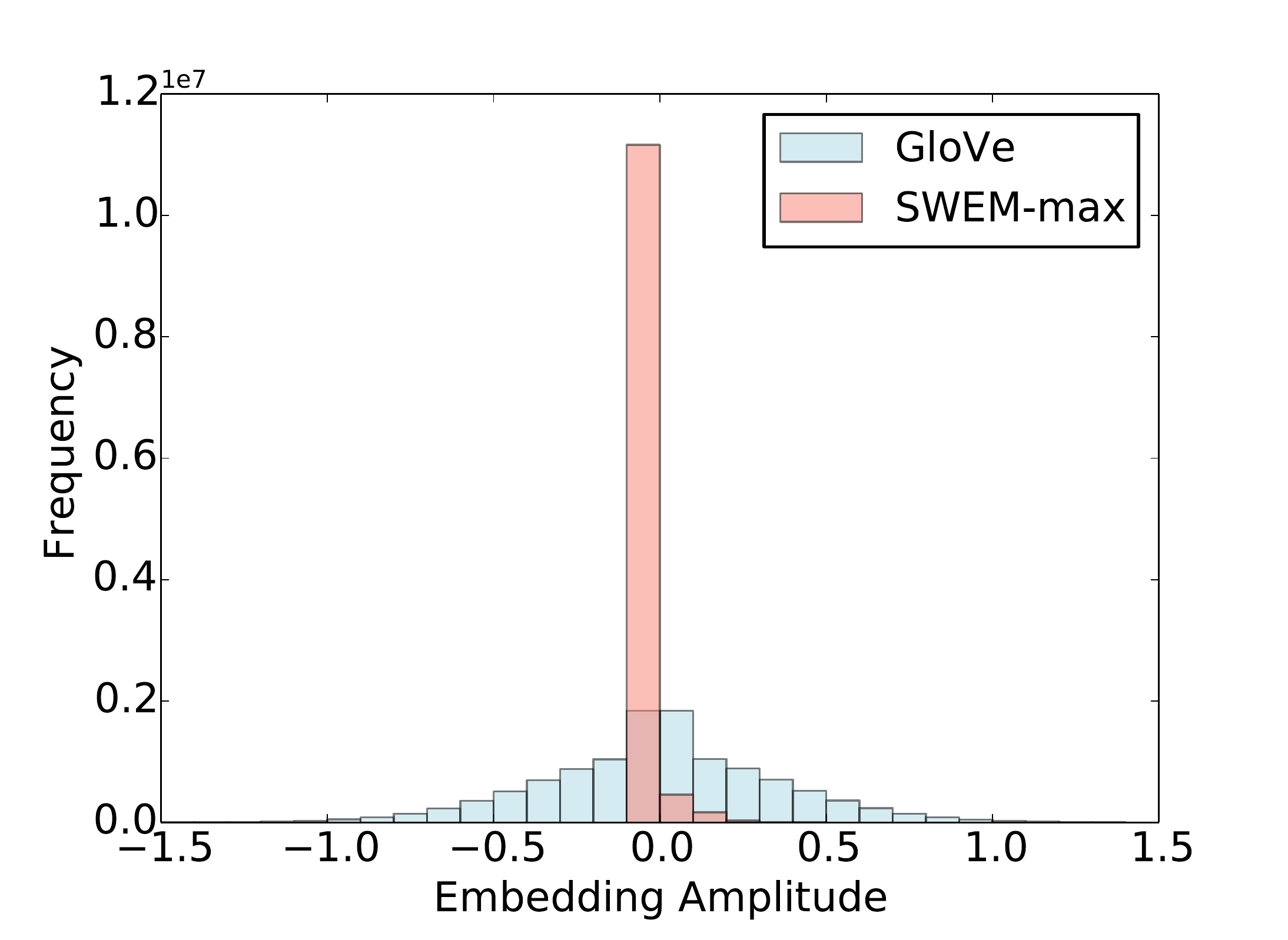}
	\vspace{-2mm}
	\captionof{figure}{Histograms for learned word embeddings (randomly initialized) of SWEM-\emph{max} and GloVe embeddings for the same vocabulary, trained on the Yahoo! Answer dataset.}
	\label{fig:sparsity}
	\vspace{-2mm}
\end{figure}

Although the proposed SWEM-\emph{max} variant generally performs a slightly worse than SWEM-\emph{aver}, it extracts complementary features from SWEM-\emph{aver}, and hence in most cases SWEM-\emph{concat} exhibits the best performance among all SWEM variants.
More importantly, we found that the word embeddings learned from  SWEM-\emph{max} tend to be sparse.
We trained our SWEM-\emph{max} model on the Yahoo datasets (randomly initialized). With the learned embeddings, we plot the values for each of the word embedding dimensions, for the entire vocabulary.
As shown in Figure~\ref{fig:sparsity}, most of the values are highly concentrated around zero, indicating that the word embeddings learned are very sparse. On the contrary, the GloVe word embeddings, for the same vocabulary, are considerably denser than the embeddings learned from SWEM-\emph{max}.
This suggests that the model may only depend on a few key words, among the entire vocabulary, for predictions (since most words do not contribute to the max-pooling operation in SWEM-\emph{max}). Through the embedding, the model learns the important words for a given task (those words with non-zero embedding components). \par 

\begin{table*}[t!] 
	\vspace{0mm}
	\centering
	\def\arraystretch{1.0}
	\begin{small}
		\begin{tabular}{c||c|cc|cc|c|cc}
			\toprule[1.2pt]
			& &\multicolumn{2}{c|}{\textbf{MultiNLI}}  &  &  &   \\
			\textbf{Model} & \textbf{SNLI} & \textbf{Matched} & \textbf{Mismatched} & \multicolumn{2}{c|}{\textbf{WikiQA}} & \textbf{Quora} &  	\multicolumn{2}{c}{\textbf{MSRP}} \\
			\hline
			& \emph{Acc.} &\emph{Acc.} & \emph{Acc.}  & \emph{MAP} & \emph{MRR}   & \emph{Acc.} & \emph{Acc.} & \emph{F1} \\
			\hline
			CNN    & 82.1 & 65.0 & 65.3 & 0.6752 & 0.6890 &  79.60 & 69.9 & 80.9  \\
			LSTM   & 80.6 &  66.9$^{\ast}$ & 66.9$^{\ast}$ &  \bf{0.6820} & \bf{0.6988} &  82.58 & 70.6 & 80.5 \\
			\hline
			SWEM-\emph{aver}         & 82.3 & 66.5 & 66.2 &  \bf{0.6808} & \bf{0.6922} &  82.68 &   71.0 & 81.1  \\	
			SWEM-\emph{max}         & \bf{83.8}  & \textbf{68.2} & \textbf{67.7} & 0.6613 & 0.6717 & 82.20 & 70.6 & 80.8  \\	
			SWEM-\emph{concat}         & 83.3 & 67.9 & 67.6 & 0.6788 & 0.6908 &  \bf{83.03} & \bf{71.5} & \bf{81.3} \\	
			\bottomrule[1.2pt]
		\end{tabular}
		\caption{Performance of different models on matching natural language sentences. Results with $^{\ast}$ are for Bidirectional LSTM, reported in \citet{williams2017broad}. Our reported results on MultiNLI are only trained MultiNLI training set (without training data from SNLI). For MSRP dataset, we follow the setup in \citet{hu2014convolutional} and do not use any additional features.}
		\label{tab:matching}
	\end{small}
	\vspace{-3mm}
\end{table*}

In this regard, the nature of max-pooling process gives rise to a more interpretable model.
For a document, only the word with largest value in each embedding dimension is employed for the final representation.
Thus, we suspect that semantically similar words may have large values in some shared dimensions.
So motivated, after training the SWEM-\emph{max} model on the Yahoo dataset, we selected five words with the largest values, among the entire vocabulary, for each word embedding dimension (these words are selected preferentially in the corresponding dimension, by the max operation).
As shown in Table~\ref{tab:similar}, the words chosen wrt each embedding dimension are indeed highly relevant and correspond to a common topic (the topics are inferred from words).
For example, the words in the first column of Table~\ref{tab:similar} are all political terms, which could be assigned to the \emph{Politics \& Government} topic.
Note that our model can even learn locally interpretable structure that is not explicitly indicated by the label information.
For instance, all words in the fifth column are \emph{Chemistry}-related.
However, we do not have a chemistry label in the dataset, and regardless they should belong to the \emph{Science} topic.


\subsection{Text Sequence Matching}
To gain a deeper understanding regarding the modeling capacity of word embeddings, we further investigate the problem of sentence matching, including natural language inference, answer sentence selection and paraphrase identification.
The corresponding performance metrics are shown in Table~\ref{tab:matching}.
Surprisingly, on most of the datasets considered (except WikiQA), SWEM demonstrates the best results compared with those with CNN or the LSTM encoder.
Notably, on SNLI dataset, we observe that SWEM-\emph{max} performs the best among all SWEM variants, consistent with the findings in \citet{nie2017shortcut, conneau2017supervised}, that \emph{max-pooling} over BiLSTM hidden units outperforms average pooling operation on SNLI dataset.
As a result, with only 120K parameters, our SWEM-\emph{max} achieves a test accuracy of 83.8\%, which is very competitive among state-of-the-art sentence encoding-based models (in terms of both performance and number of parameters)\footnote{See leaderboard at \url{https://nlp.stanford.edu/projects/snli/} for details.}.

The strong results of the SWEM approach on these tasks may stem from the fact that when matching natural language sentences, it is sufficient in most cases to simply model the word-level alignments between two sequences \citep{parikh2016decomposable}.
From this perspective, word-order information becomes much less useful for predicting relationship between sentences.
Moreover, considering the simpler model architecture of SWEM, they could be much easier to be optimized than LSTM or CNN-based models, and thus give rise to better empirical results.

\subsubsection{Importance of word-order information}\label{important}
One possible disadvantage of SWEM is that it ignores the word-order information within a text sequence, which could be potentially captured by CNN- or LSTM-based models.
However, we empirically found that except for sentiment analysis, SWEM exhibits similar or even superior performance as the CNN or LSTM on a variety of tasks.
In this regard, one natural question would be: how important are word-order features for these tasks?
To this end, we randomly shuffle the words for every sentence in the training set, while keeping the original word order for samples in the test set.
The motivation here is to remove the word-order features from the training set and examine how sensitive the performance on different tasks are to word-order information.
We use LSTM as the model for this purpose since it can captures word-order information from the original training set.

\begin{table} [h!]
	\centering 
	\vspace{-2mm}
	\def\arraystretch{1.2}
	\begin{small}
		\begin{tabular}{c||c|c|c}
			\toprule[1.2pt]
			\textbf{Datasets} &  \textbf{Yahoo}  & \textbf{Yelp P.} & \textbf{SNLI} \\
			\hline
			\textbf{Original} & 72.78  & 95.11 & 78.02  \\
			\textbf{Shuffled} & 72.89 & 93.49  & 77.68  \\
			\bottomrule[1.2pt]
		\end{tabular}
	\end{small}
	\caption{Test accuracy for LSTM model trained on original/shuffled training set.}
	\label{tab:order}
	\vspace{-2mm}
\end{table}

The results on three distinct tasks are shown in Table~\ref{tab:order}.
Somewhat surprisingly, for Yahoo and SNLI datasets, the LSTM model trained on shuffled training set shows comparable accuracies to those trained on the original dataset, indicating that word-order information does not contribute significantly on these two problems, \emph{i.e.}, topic categorization and textual entailment.
However, on the Yelp polarity dataset, the results drop noticeably, further suggesting that word-order does matter for sentiment analysis (as indicated above from a different perspective). 

Notably, the performance of LSTM on the Yelp dataset with a shuffled training set is very close to our results with SWEM, indicating that the main difference between LSTM and SWEM may be due to the ability of the former to capture word-order features. Both observations are in consistent with our experimental results in the previous section.

\begin{table}[t!]
	\centering
	\resizebox{\columnwidth}{!}{%
		\begin{tabular}{l p{2.6in}}
			\toprule[1.2pt]
			\textbf{Negative}: & Friendly staff and nice selection of vegetarian options. Food \textbf{\textcolor{blue}{is just okay}}, \textbf{\textcolor{blue}{not great}}. \textbf{\textcolor{blue}{Makes me wonder why everyone likes}} food fight so much.   \\
			\hline
			\hline
			\textbf{Positive}: & The store is small, but it carries specialties that are difficult to find in Pittsburgh. I \textbf{\textcolor{blue}{was particularly excited}} to find middle eastern chili sauce and chocolate covered turkish delights.\\
			\bottomrule[1.2pt]
		\end{tabular}
	}
	\caption{Test samples from Yelp Polarity dataset for which LSTM gives wrong predictions with shuffled training data, but predicts correctly with the original training set.}
	\label{tab:order_matter}
	\vspace{-4mm}
\end{table}

\paragraph{Case Study}
To understand what type of sentences are sensitive to word-order information, we further show those samples that are wrongly predicted because of the shuffling of training data in Table~\ref{tab:order_matter}.
Taking the first sentence as an example, several words in the review are generally positive, \emph{i.e.} \emph{friendly}, \emph{nice}, \emph{okay}, \emph{great} and \emph{likes}.
However, the most vital features for predicting the sentiment of this sentence could be the phrase/sentence \emph{`is just okay'}, \emph{`not great'} or \emph{`makes me wonder why everyone likes'}, which cannot be captured without considering word-order features. 
It is worth noting the hints for predictions in this case are actually $n$-gram phrases from the input document. 

\subsection{SWEM-\emph{hier} for sentiment analysis} \label{LG}
As demonstrated in Section~\ref{important}, word-order information plays a vital role for sentiment analysis tasks.
However, according to the case study above, the most important features for sentiment prediction may be some key $n$-gram phrase/words from the input document.
We hypothesize that incorporating information about the local word-order, \emph{i.e.}, $n$-gram features, is likely to 
largely mitigate the limitations of the above three SWEM variants.
Inspired by this observation, we propose using another simple pooling operation termed as hierarchical (SWEM-\emph{hier}), as detailed in Section~\ref{swem_lg}. 
We evaluate this method on the two document-level sentiment analysis tasks and the results are shown in the last row of Table~\ref{tab:document}.

SWEM-\emph{hier} greatly outperforms the other three SWEM variants, and the corresponding accuracies are comparable to the results of CNN or LSTM (Table~\ref{tab:document}).
This indicates that the proposed hierarchical pooling operation manages to abstract spatial (word-order) information from the input sequence, which is beneficial for performance in sentiment analysis tasks.

\begin{table*}[!ht] 
	\vspace{-2mm}
	\centering
	\def\arraystretch{1.0}
	\begin{small}
		\begin{tabular}{c||c|c|c|c|c}
			\toprule[1.2pt]
			\textbf{Model} &  \textbf{MR} & \textbf{SST-1} & \textbf{SST-2} &  	\textbf{Subj} &  	\textbf{TREC} \\
			\hline
			RAE \cite{socher2011semi}      & 77.7 & 43.2 & 82.4 & -- & --    \\
			MV-RNN \cite{socher2012semantic}  & 79.0 & 44.4& 82.9 & -- & --  \\
			LSTM \cite{tai2015improved}   & -- & 46.4 & 84.9 & -- & --   \\
			RNN \cite{zhao2015self}   & 77.2 & -- & -- & \bf{93.7} & 90.2    \\
			Constituency Tree-LSTM \cite{tai2015improved}   & - & \bf{51.0} & 88.0 & - & -   \\
			Dynamic CNN \cite{kalchbrenner2014convolutional}  & -- & 48.5 & 86.8 & -- & 93.0  \\
			CNN \cite{kim2014convolutional}     & \bf{81.5} & 48.0 & \bf{88.1} & 93.4 & \bf{93.6}   \\
			DAN-ROOT \cite{iyyer2015deep}   & - & 46.9 & 85.7 & - & -   \\
			\hline
			SWEM-\emph{aver}          & 77.6 &45.2 & 83.9 & 92.5 & \bf{92.2}  \\
			SWEM-\emph{max}         & 76.9 &44.1 & 83.6 & 91.2 & 89.0 \\
			SWEM-\emph{concat}         & \bf{78.2} &\bf{46.1} & \bf{84.3} & \bf{93.0} & 91.8 \\
			\bottomrule[1.2pt]
		\end{tabular}
	\end{small}
	\vspace{-2mm}
	\caption{Test accuracies with different compositional functions on (short) sentence classifications.}
	\label{tab:sentence}
	\vspace{-2mm}
\end{table*}

\subsection{Short Sentence Processing} \label{short}
%
We now consider sentence-classification tasks (with approximately 20 words on average).
We experiment on three sentiment classification datasets, \emph{i.e.}, MR, SST-1, SST-2, as well as subjectivity classification (Subj) and question classification (TREC).
The corresponding results are shown in Table~\ref{tab:sentence}.
Compared with CNN/LSTM compositional functions, SWEM yields inferior accuracies on sentiment analysis datasets, consistent with our observation in the case of document categorization.
However, SWEM exhibits comparable performance on the other two tasks, again with much less parameters and faster training.
Further, we investigate two sequence tagging tasks: the standard CoNLL2000 chunking and CoNLL2003 NER datasets.
Results are shown in the Supplementary Material, where LSTM and CNN again perform better than SWEMs.
Generally, SWEM is less effective at extracting representations from \emph{short} sentences than from \emph{long} documents.
This may be due to the fact that for a shorter text sequence, word-order features tend to be more important since the semantic information provided by word embeddings alone is relatively limited.

Moreover, we note that the results on these relatively small datasets are highly sensitive to model regularization techniques due to the overfitting issues.
In this regard, one interesting future direction may be to develop specific regularization strategies for the SWEM framework, and thus make them work better on small sentence classification datasets.

\section{Discussion}

\subsection{Comparison via subspace training}

\begin{figure}[t!] \centering
	\vspace{-0mm}
	\begin{tabular}{c c}		
		\hspace{-6mm}
		\includegraphics[width=3.80cm]{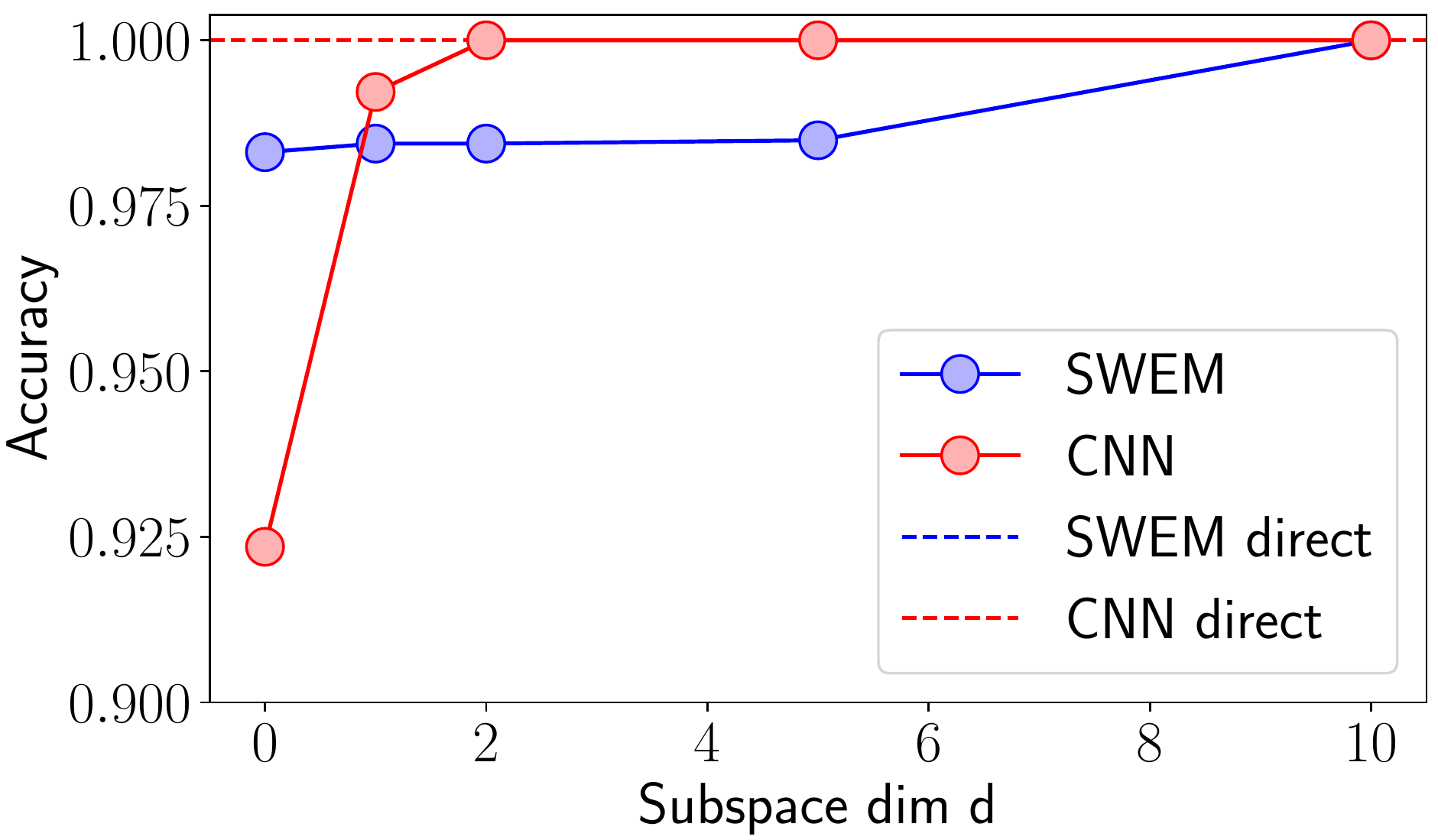}  
		& 
		\hspace{-6mm}
		\includegraphics[width=3.80cm]{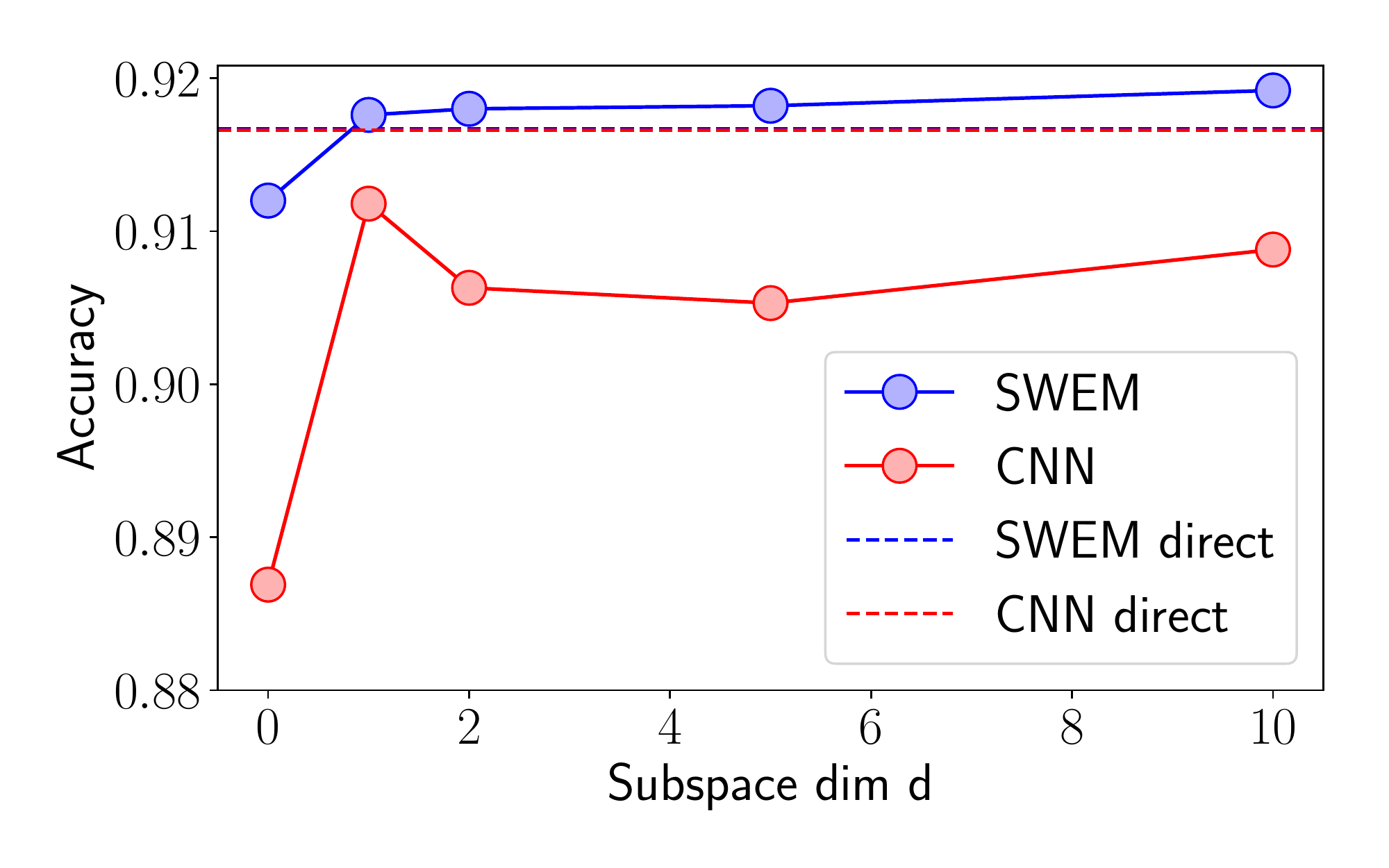}  \\
			\hspace{-4mm}
		(a) Training on AG News \hspace{-0mm} & \hspace{-5mm}
		(b) Testing on AG News\\
		\hspace{-6mm}
		\includegraphics[width=3.80cm]{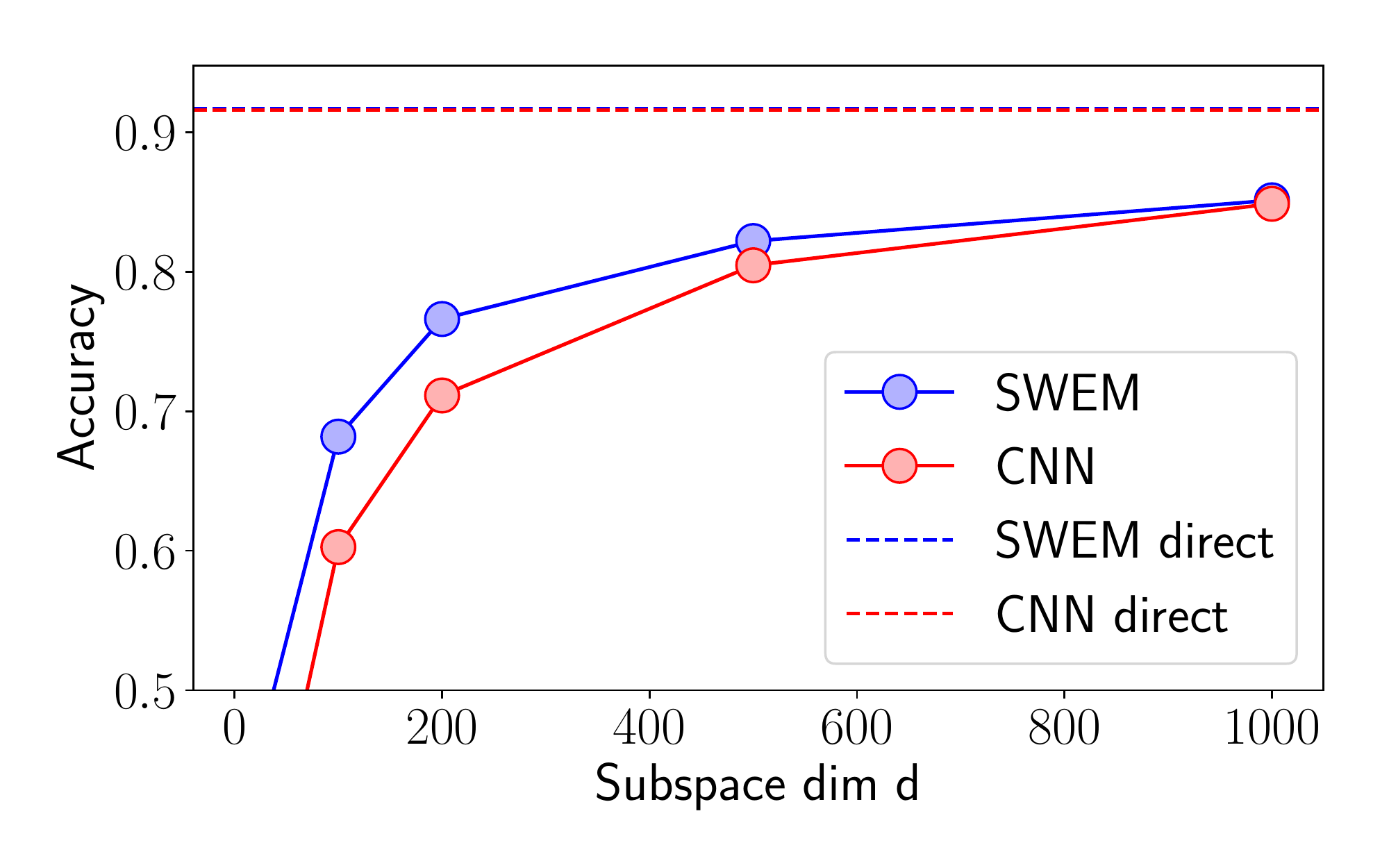}  
		& 
		\hspace{-6mm}
		\includegraphics[width=3.80cm]{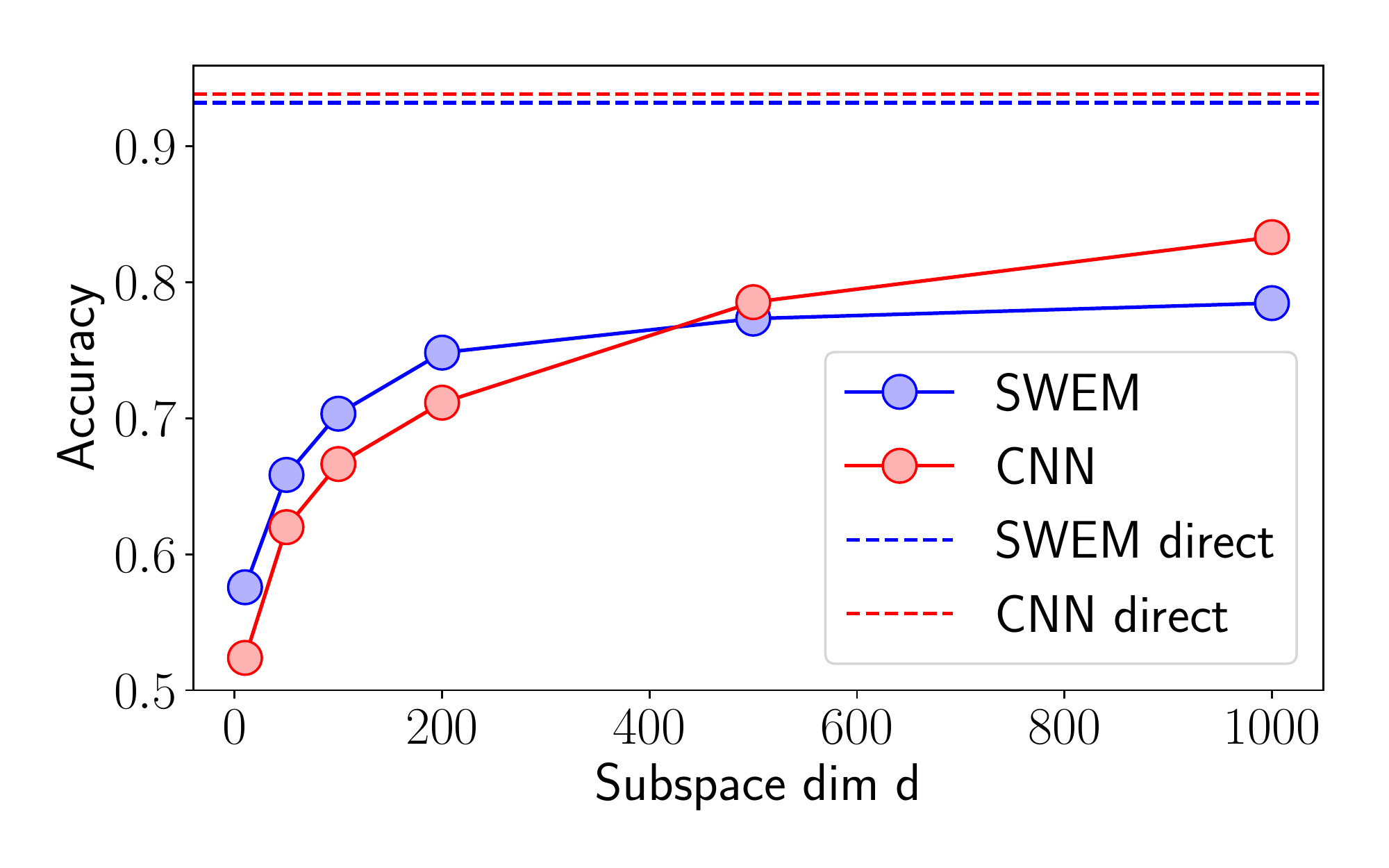}  \\
		\hspace{-4mm}
		(c) Testing on AG News \hspace{-0mm} & \hspace{-5mm}
		(d)Testing on Yelp P.
	\end{tabular}
	\vspace{-3mm}
	\caption{Performance of subspace training. Word embeddings are optimized in (a)(b), and frozen in (c)(d).}
	\vspace{-6mm}
	\label{fig:subspace}
\end{figure}

We use {\it subspace training}~\cite{li_id_2018_ICLR}  to measure the model complexity in text classification problems. It constrains the optimization of the trainable parameters in a subspace of low dimension $d$, the intrinsic dimension $d_{\rm int}$ defines the minimum $d$ that yield a good solution. Two models are studied: the SWEM-\emph{max} variant, and the CNN model including a convolutional layer followed by a FC layer.
We consider two settings:

(1) The word embeddings are randomly intialized, and optimized jointly with the model parameters. We show the performance of direct and subspace training on AG News dataset in Figure~\ref{fig:subspace} (a)(b). The two models trained via direct method share almost identical perfomrnace on training and testing. The subspace training yields similar accuracy with direct training for very small $d$, even when model parameters are not trained at all ($d=0$). This is because the word embeddings have the full degrees of freedom to adjust to achieve good solutions, regardless of the employed models. SWEM seems to have an easier loss landspace than CNN for word embeddings to find the best solutions. According to Occam's razor, simple models are preferred, if all else are the same.

(2) The pre-trained GloVe are frozen for the word embeddings, and only the model parameters are optimized. The results on testing datasets of AG News and Yelp P. are shown in Figure~\ref{fig:subspace} (c)(d), respectively. SWEM shows significantly higher accuracy than CNN for a large range of low subspace dimension, indicating that SWEM is more parameter-efficient to get a decent solution. In Figure~\ref{fig:subspace}(c), if we set the performance threshold as 80\% testing accuracy, SWEM exhibits a lower $d_{\rm int}$ than CNN on AG News dataset. However, in Figure~\ref{fig:subspace}(d), CNN can leverage more trainable parameters to achieve higher accuracy when $d$ is large.

\subsection{Linear classifiers}
\vspace{-1mm}
To further investigate the quality of representations learned from SWEMs, we employ a linear classifier on top of the representations for prediction, instead of a non-linear MLP layer as in the previous section. It turned out that utilizing a linear classifier only leads to a very small performance drop for both Yahoo! Ans. (from $73.53\%$ to $73.18\%$) and Yelp P. datasets (from $93.76\%$ to $93.66\%$) . This observation highlights that SWEMs are able to extract robust and informative sentence representations despite their simplicity.

\subsection{Extension to other languages}
\vspace{-1mm}
We have also tried our SWEM-concat and SWEM-hier models on Sogou news corpus (with the same experimental setup as \cite{zhang2015character}), which is a \emph{Chinese} dataset represented by Pinyin (a phonetic romanization of Chinese). SWEM-concat yields an accuracy of $91.3\%$, while SWEM-hier (with a local window size of 5) obtains an accuracy of $96.2\%$ on the test set. Notably, the performance of SWEM-hier is comparable to the best accuracies of CNN ($95.6\%$) and LSTM ($95.2\%$), as reported in \cite{zhang2015character}. This indicates that hierarchical pooling is more suitable than average/max pooling for Chinese text classification, by taking spatial information into account. It also implies that Chinese is more sensitive to local word-order features than English.

\vspace{-1mm}
\section{Conclusions}
\vspace{-2mm}
We have performed a comparative study between SWEM (with parameter-free pooling operations) and CNN or LSTM-based models, to represent text sequences on 17 NLP datasets.
We further validated our experimental findings through additional exploration, and revealed some general rules for rationally selecting compositional functions for distinct problems.
Our findings regarding when (and why) simple pooling operations are enough for text sequence representations are summarized as follows: \nocite{shen2017adaptive} \par 
\smallskip 
\noindent$\bullet$ Simple pooling operations are surprisingly effective at representing longer documents (with hundreds of words), while recurrent/convolutional compositional functions are most effective when constructing representations for short sentences.\par
\smallskip 
\noindent$\bullet$ Sentiment analysis tasks are more sensitive to word-order features than topic categorization tasks. However, a simple \emph{hierarchical pooling layer} proposed here achieves comparable results to LSTM/CNN on sentiment analysis tasks.  \par
\smallskip 
\noindent$\bullet$ To match natural language sentences, \emph{e.g.}, textual entailment, answer sentence selection, \emph{etc.}, simple pooling operations already exhibit similar or even superior results, compared to CNN and LSTM.\par
\smallskip 
\noindent$\bullet$ In SWEM with max-pooling operation, each \emph{individual dimension} of the word embeddings contains interpretable semantic patterns, and groups together words with a common theme or \emph{topic}.\par
\smallskip 

%

\bibliography{word_emb}
\bibliographystyle{acl_natbib}

\clearpage

\section*{Appendix I: Experimental Setup}

\subsection{Data statistics}
We consider a wide range of text-representation-based tasks in this paper, including \emph{document categorization}, \emph{text sequence matching} and \emph{(short) sentence classification}. For document classification tasks, we use the same data splits in \cite{zhang2015character} (downloaded from https://goo.gl/QaRpr7); for short sentence classification, we employ the same training/testing data and preprocessing procedure with \cite{kim2014convolutional}. The statistics and corresponding types of these datasets are summarized in Table~\ref{tab:statistics}
\begin{table}[ht!] 
	\centering
	\def\arraystretch{1.2}
	\vspace{-0mm}
	\resizebox{\columnwidth}{!}{%
		\begin{tabular}{c||c|c|c|c}
			\toprule[1.2pt]
			\textbf{Datasets} &  	\textbf{\#w} & \textbf{\#c} & 	\textbf{Train} & \textbf{Types}  \\
			\hline
			Yahoo        & 104 & 10 & 1,400K& Topic categorization    \\
			AG News        & 43 & 4 & 120K & Topic categorization    \\
			Yelp P.       & 138 & 2 & 560K & Sentiment analysis     \\
			Yelp F.      & 152 & 5 & 650K & Sentiment analysis    \\
			DBpedia       & 57 & 14 & 560K & Ontology classification  \\
			\hline
			SNLI         & 11 / 6 & 3 & 549K & Textual Entailment     \\
			MultiNLI           & 21/11  & 3 & 393K & Textual Entailment     \\
			WikiQA       & 7 / 26 & 2 & 20K & Question answering   \\
			Quora         & 13 / 13 & 2 & 384K & Paraphrase identification    \\
			MSRP        & 23 / 23 & 2 & 4K & Paraphrase identification   \\
			\hline
			MR        & 20 & 2 & 11K & Sentiment analysis    \\
			SST-1        & 18 & 5 & 12K & Sentiment analysis    \\
			SST-2        & 19 & 2 & 10K & Sentiment analysis     \\
			Subj        & 23 & 2 & 10K & Subjectivity classification   \\
			TREC        & 10 & 6 & 6K & Question classification   \\
			\bottomrule[1.2pt]
		\end{tabular}
	}
	\caption{Data Statistics. Where $\textbf{\#w}$, $\textbf{\#c}$ and Train denote the average number of words, the number of classes and the size of training set, respectively. For sentence matching datasets, $\textbf{\#w}$ stands for the average length for the two corresponding sentences.}
	\label{tab:statistics}
\end{table}

\subsection{Sequence Tagging Results}

\begin{table} [!h]
	\centering 
	\def\arraystretch{1.2}
	\begin{small}
		\begin{tabular}{c||c|c}
			\toprule[1.2pt]
			\textbf{Datasets} &  \textbf{CoNLL2000}  & \textbf{CoNLL2003}  \\
			\hline
			 \textbf{CNN-CRF} & 94.32  & 89.59 \\
			\textbf{BI-LSTM-CRF}  &  94.46  &   90.10  \\
			\textbf{SWEM-CRF}  & 90.34&   86.28 \\
			\bottomrule[1.2pt]
		\end{tabular}
	\end{small}
	\caption{The results (F1 score) on sequence tagging tasks. }
	\label{tab:tagging}
\end{table}

SWEM-CRF indicates that CRF is directly operated on top of the word embedding layer and make predictions for each word (there is no contextual/word-order information before CRF layer, compared to CNN-CRF or BI-LSTM-CRF). As shown above, CNN-CRF and BI-LSTM-CRF consistently outperform SWEM-CRF on both sequence tagging tasks, although the training takes around 4 to 5 times longer (for BI-LSTM-CRF) than SWEM-CRF. This suggests that for chunking and NER, compositional functions such as LSTM or CNN are very necessary, because of the sequential (order-sensitive) nature of sequence tagging tasks. 

\subsection{What are the key words used for predictions?}
\begin{figure}[!h]
	\vspace{-1mm}
	\centering
	\includegraphics[scale=0.2]{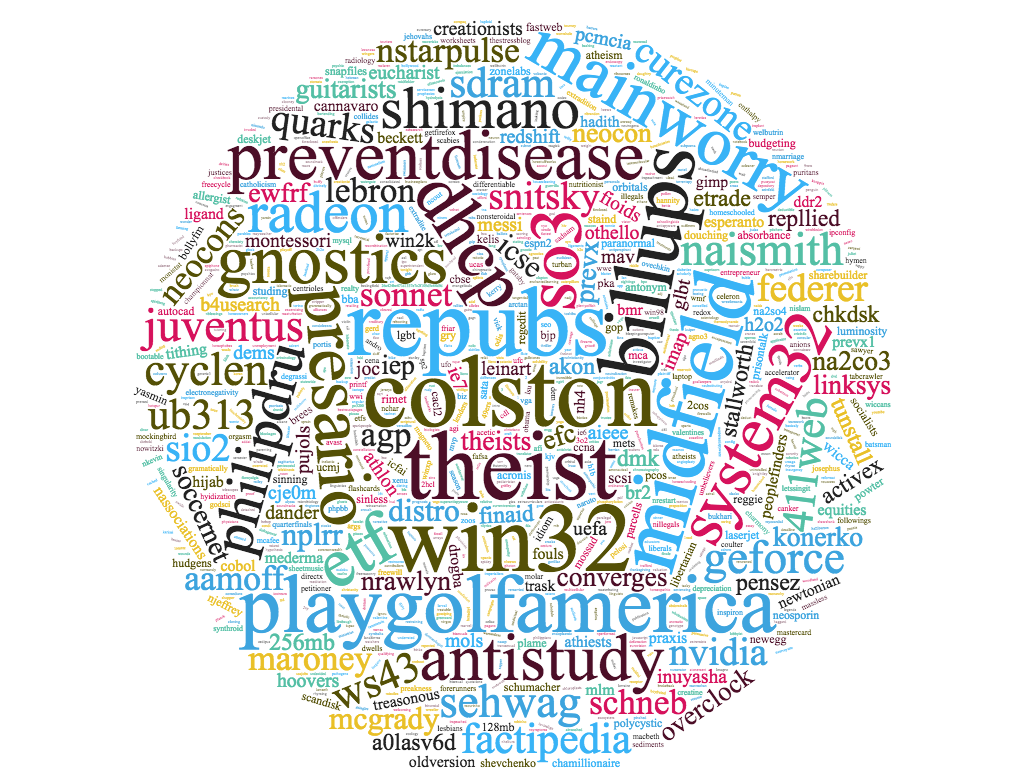}
	\vspace{0mm}
	\captionof{figure}{The top 10 words for each word embeddings' dimension. }
	\label{fig:key}
	\vspace{-1mm}
\end{figure}

Given the sparsity of word embeddings, one natural question would be: What are those \emph{key words} that are leveraged by the model to make predictions? To this end, after training SWEM-\emph{max} on Yahoo! Answer dataset, we selected the top-10 words (with the maximum values in that dimension) for every word embedding dimension. The results are visualized in Figure~\ref{fig:key}. These words are indeed very predictive since they are likely to occur in documents with a specific topic, as discussed above. Another interesting observation is that the frequencies of these words are actually quite low in the training set (\emph{e.g.} \emph{colston}: 320, \emph{repubs}: 255 \emph{win32}: 276), considering the large size of the training set (1,400K). This suggests that the model is utilizing those relatively rare, yet representative words of each topic for the final predictions.

\subsection{How many word embedding dimensions are needed?} \label{need}
Since there are no compositional parameters in SWEM, the component that contains the semantic information of a text sequence is the word embedding.
Thus, it is of interest to see how many word embedding dimensions are needed for a SWEM architecture to perform well.
To this end, we vary the dimension from 3 to 1000 and train a SWEM-\emph{concat} model on the Yahoo dataset.
For fair comparison, the word embeddings are randomly initialized in this experiment, since there are no pre-trained word vectors, such as GloVe \citep{pennington2014glove}, for some dimensions we consider.
As shown in Table~\ref{tab:dim}, the model exhibits higher accuracy with larger word embedding dimensions. 
This is not surprising since with more embedding dimensions, more semantic features could be potentially encapsulated.
However, we also observe that even with only 10 dimensions, SWEM demonstrates comparable results relative to the case with 1000 dimensions, suggesting that word embeddings are very efficient at abstracting semantic information into fixed-length vectors.
This property indicates that we may further reduce the number of model parameters with lower-dimensional word embeddings, while still achieving competitive results.

\subsection{Sensitivity of compositional functions to sample size} \label{semi}
\begin{figure*}
	\centering
	\includegraphics[scale=0.53]{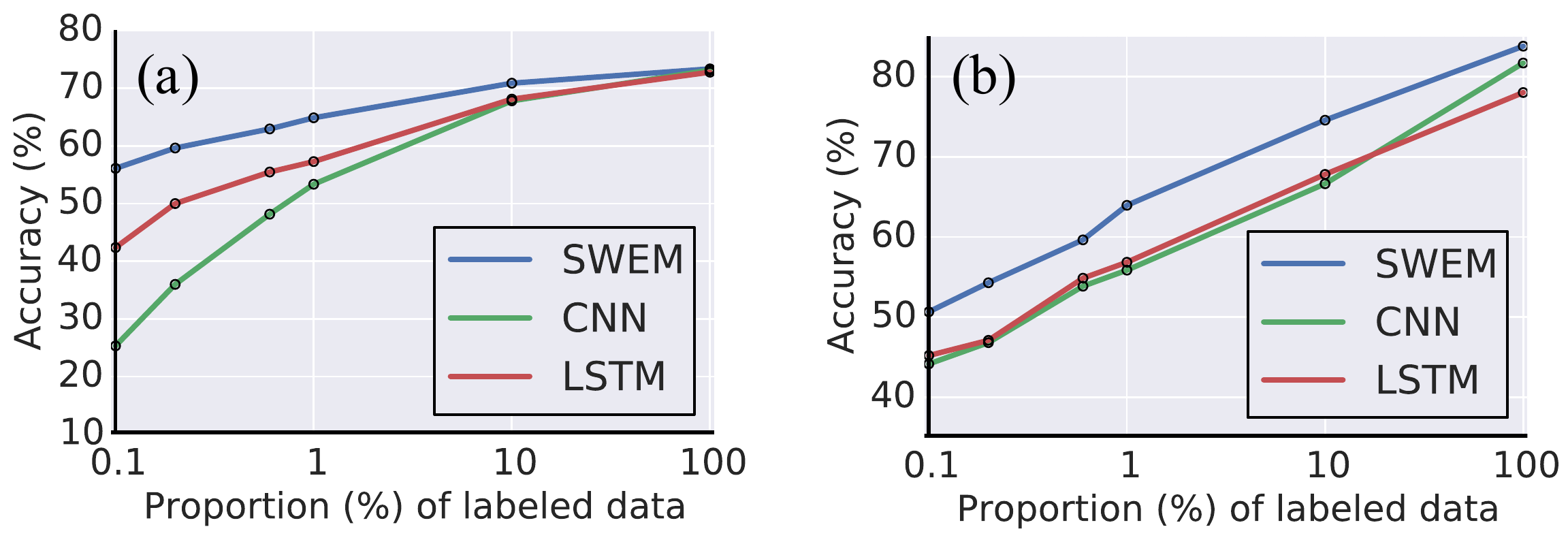}
	\captionof{figure}{The test accuracy comparisons between SWEM and CNN/LSTM models on (a) Yahoo! Answers dataset and (b) SNLI dataset, with different proportions of training data (ranging from $0.1\%$ to $100\%$). }
	\label{fig:partial}
\end{figure*}

To explore the robustness of different compositional functions, we consider another application scenario, where we only have a limited number of training data, \emph{e.g.}, when labeled data are expensive to obtain.
To investigate this, we re-run the experiments on Yahoo and SNLI datasets, while employing increasing proportions of the original training set.
Specifically, we use $0.1\%$, $0.2\%$,  $0.6\%$, $1.0\%$, $10\%$, $100\%$ for comparison; the corresponding results are shown in Figure~\ref{fig:partial}.

\begin{table}
	\centering
	\def\arraystretch{1.2}
	\resizebox{\columnwidth}{!}{%
		\begin{tabular}{c||c|c|c|c|c|c}
			
			\toprule[1.2pt]
			\textbf{\# Dim.} &  $3$  & $10$ & $30$ & $100$ & $300$ & $1000$ \\
			\hline
			\textbf{Yahoo} & 64.05  & \textbf{72.62} & 73.13 & 73.12 & 73.24 & 73.31  \\
			\bottomrule[1.2pt]
		\end{tabular}
	}
	\caption{Test accuracy of SWEM on Yahoo dataset with a wide range of word embedding dimensions.}
	\label{tab:dim}
\end{table}

Surprisingly, SWEM consistently outperforms CNN and LSTM models by a large margin, on a wide range of training data proportions.
For instance, with $0.1\%$ of the training samples from Yahoo dataset (around 1.4K labeled data), SWEM achieves an accuracy of $56.10\%$, which is much better than that of models with CNN ($25.32\%$) or LSTM ($42.37\%$).
On the SNLI dataset, we also noticed the same trend that the SWEM architecture result in much better accuracies, with a fraction of training data.
This observation indicates that overfitting issues in CNN or LSTM-based models on text data mainly stems from over-complicated compositional functions, rather than the word embedding layer.
More importantly, SWEM tends to be a far more robust model when only limited data are available for training.

\end{document}